\documentclass[letterpaper, 10 pt, journal, twoside]{IEEEtran}

\usepackage{authblk}
\usepackage{graphics} 
\usepackage{epsfig} 
\usepackage{mathptmx} 
\usepackage{times} 
\usepackage{amsmath} 
\usepackage{amssymb}  
\usepackage{cite}
\usepackage{xcolor}
\usepackage{booktabs}
\usepackage{float}
\usepackage{multirow}
\usepackage{caption}
\usepackage{newtxtext, newtxmath}
\usepackage{tabularx}
\usepackage{longtable}
\usepackage{hyperref}

\usepackage{flushend}
\usepackage{amsmath}
\usepackage{stfloats}
\usepackage{balance}
\usepackage{xspace}
\usepackage{doi}
\def\themodel{{{VR-Robo}}\xspace} 

\begin{document}

\title{\LARGE \bf
\themodel: A Real-to-Sim-to-Real Framework for Visual Robot \\ Navigation and Locomotion
}

\author{Shaoting Zhu$^{12*}$, Linzhan Mou$^{23*}$, Derun Li$^{24}$, Baijun Ye$^{13}$, Runhan Huang$^{12}$, Hang Zhao†$^{123}$\vspace{-3mm}\\
$^*$Equal contribution\quad†Corresponding author\quad
\href{https://vr-robo.github.io/}{\textbf{VR-Robo.github.io}\xspace}\vspace{-0.3in}
\thanks{Manuscript received: January 31, 2025; Revised April 14, 2025; Accepted May 10, 2025.}
\thanks{This paper was recommended for publication by
Editor Olivier Stasse upon evaluation of the Associate Editor and Reviewers’ comments.}%
\thanks{$^{1}$The authors are with Institute for Interdisciplinary Information Sciences, Tsinghua University, China. {\tt\footnotesize \{zhust24, yebj24, hrh22\}@mails.tsi\\ \tt\footnotesize nghua.edu.cn}, \tt\footnotesize 
hangzhao@tsinghua.edu.cn}%
\thanks{$^{2}$The authors are with Shanghai Qi Zhi Institute, Shanghai, China. \tt\footnotesize moulz@seas.upenn.edu}%
\thanks{$^{3}$The authors are with Galaxea AI, Beijing, China.}%
\thanks{$^{4}$Derun Li is with School of Electronics, Information and Electrical Engineering, Shanghai Jiao Tong University, China. \tt\footnotesize ldr.dylan@sjtu.edu.cn}%
\thanks{Digital Object Identifier (DOI): see top of this page.}%
}

\markboth{IEEE Robotics and Automation Letters. Preprint Version. Accepted May, 2025}
{Zhu \MakeLowercase{\textit{et al.}}: VR-Robo: A Real-to-Sim-to-Real Framework for Visual Robot Navigation and Locomotion} 

\maketitle


\begin{abstract}
Recent success in legged robot locomotion is attributed to the integration of reinforcement learning and physical simulators. 
However, these policies often encounter challenges when deployed in real-world environments due to sim-to-real gaps, as simulators typically fail to replicate visual realism and complex real-world geometry. Moreover, the lack of realistic visual rendering limits the ability of these policies to support high-level tasks requiring RGB-based perception like ego-centric navigation. This paper presents a \textit{Real-to-Sim-to-Real} framework that generates photorealistic and physically interactive ``digital twin'' simulation environments for visual navigation and locomotion learning. Our approach leverages 3D Gaussian Splatting (3DGS) based scene reconstruction from multi-view images and integrates these environments into simulations that support ego-centric visual perception and mesh-based physical interactions. To demonstrate its effectiveness, we train a reinforcement learning policy within the simulator to perform a visual goal-tracking task. Extensive experiments show that our framework achieves RGB-only sim-to-real policy transfer. Additionally, our framework facilitates the rapid adaptation of robot policies with effective exploration capability in complex new environments, highlighting its potential for applications in households and factories.

\end{abstract}



\section{Introduction}


Exploring, perceiving, and interacting with the physical real world is crucial for legged robotics, which supports many applications such as household service and industrial automation. However, conducting real-world experiments poses considerable challenges due to safety risks and efficiency constraints. Consequently, training in simulation\cite{mujoco,Isaac,habitat} has emerged as an effective alternative, allowing robots to experience diverse environmental conditions, safely explore failure cases, and learn directly from their actions. Despite these advantages, transferring policies learned in simulated environments to real world remains challenging, owing to the significant \textit{Sim-to-Real} reality gap\cite{maniskill3,mujoco}.

Substantial efforts have been made to narrow the \textit{Sim-to-Real} gap in legged locomotion. Previous studies have employed cross-domain depth images to train agents in simulation, achieving impressive zero-shot policy transfer for both quadruped\cite{parkour,cheng2024extreme,hoeller2024anymal,wu2023learning} and humanoid\cite{humanoid} robots. To further integrate RGB perception into the \textit{Sim-to-Real} pipeline, LucidSim\cite{lucidsim} leverages generative models within simulation\cite{mujoco} for visual parkour. However, these depth-based and generation-based visual policies are predominantly constrained to low-level locomotion tasks, primarily because standard simulators struggle to replicate the visual fidelity and complex geometry of real-world environments.

Recently, advanced neural reconstruction techniques such as Neural Radiance Fields (NeRF)\cite{nerf} and 3D Gaussian Splatting (3DGS)\cite{3dgs} have emerged as promising solutions for creating \textit{Real-to-Sim} ``digital twins'' from real-world data. Nonetheless, most existing approaches\cite{nerf2real,robogs,robogsim,rl-gsbrige,vid2sim} focus on enhancing photorealism, offering limited support for physical interaction with complex terrains. Moreover, the simulations in these works often lack mechanisms for robust environment exploration, thus constraining their deployment in complex real-world scenarios that demand dynamic interaction.

\begin{figure}[t]
    \centering
    \includegraphics[width=1.0\linewidth]{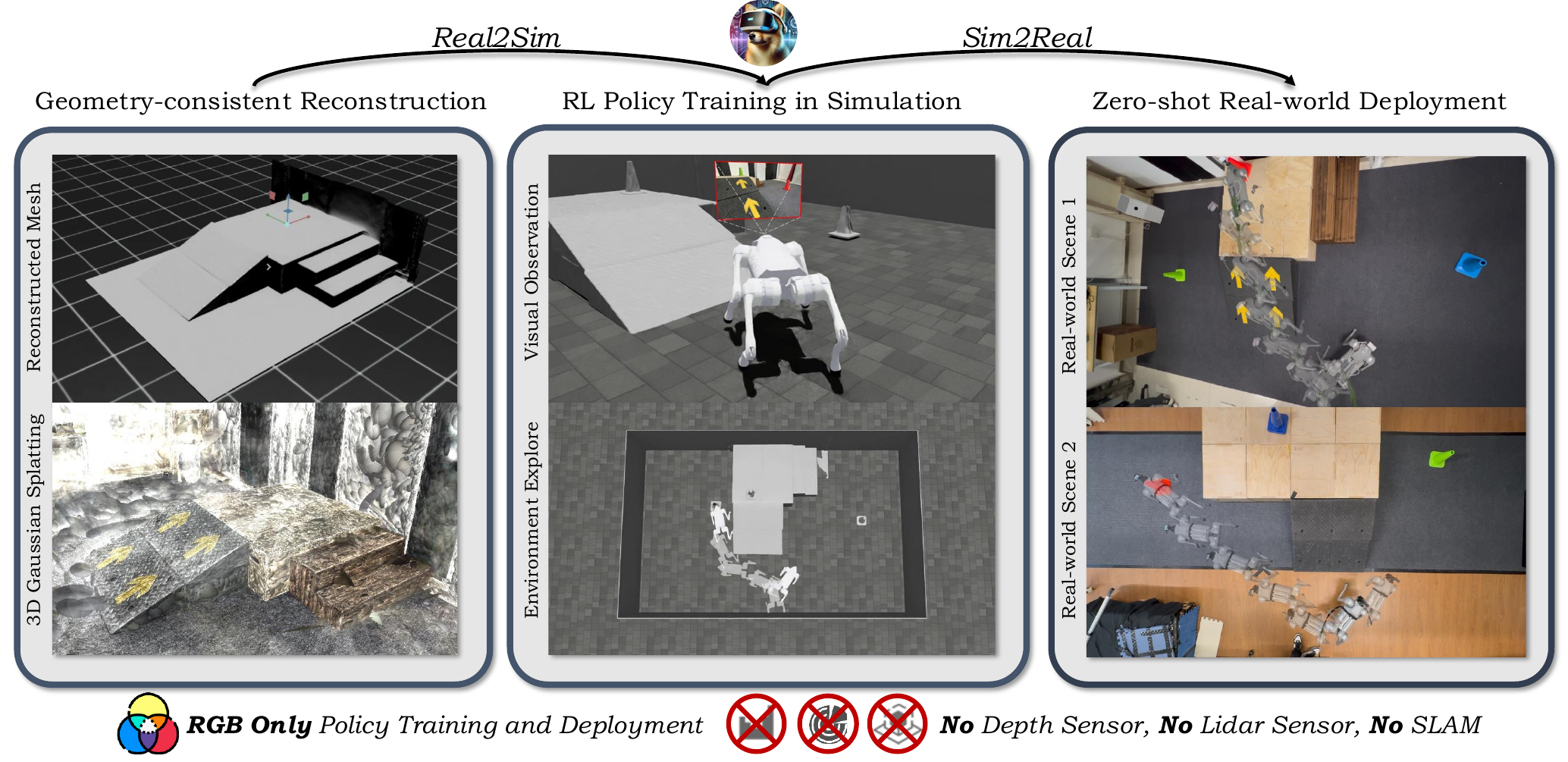}
    \vspace{-5mm}
    \caption{\textbf{Our \textit{{\themodel}} introduces a unified \textit{real-to-sim-to-real} framework.} We generate photorealistic and physically interactive ``digital twin'' simulation environments for visual policy training, and enable zero-shot real-world deployment.}
    \vspace{-6mm}
    \label{fig:teaser}
\end{figure}
\label{sec:intro}
To address these challenges, we introduce \themodel, a novel \textit{Real-to-Sim-to-Real} framework that enables realistic, interactive \textit{Real-to-Sim} simulation and reinforcement learning (RL) policy training for legged robot navigation and locomotion. Given the multi-view images, we employ 3DGS\cite{3dgs} and foundation-model priors\cite{da2} to reconstruct geometry-consistent scenes and objects. We then propose a GS-mesh hybrid representation with coordinate alignment to create a ``digital twin'' simulation environment in Isaac Sim, which supports ego-centric visual perception and mesh-based physical interactions. To enable robust policy training, we introduce an agent-object randomization and occlusion-aware scene composition strategy, as illustrated in \autoref{fig:random}.


We summarize our contributions as follows:
\begin{enumerate}
    \item \textbf{A \textit{Real-to-Sim-to-Real} framework for robot navigation and locomotion.} We propose to reduce the \textit{Sim-to-Real} gap by training RL policies within a realistic and interactive simulation environment reconstructed only from RGB images captured from real-world.
    \item \textbf{Photorealistic and physically interactive \textit{Real-to-Sim} environment reconstruction.} We introduce a novel pipeline for transferring real-world environments into simulation using a GS-mesh hybrid representation with coordinate alignment. We further incorporate object randomization and occlusion-aware scene composition for robust and efficient RL policy training.
    \item \textbf{Zero-shot \textit{Sim-to-Real} RL policy transfer.} Through extensive experimental evaluations, we demonstrate that \themodel produces effective navigation and locomotion policies in complex, real-world scenarios using RGB-only observations. In addition, we have some empirical findings for RL policy training, including the generalization ability, sample complexity, and architectural choices.
\end{enumerate}

\begin{figure*}[ht]
    \centering
    \vspace{2mm}
    \includegraphics[width=1.0\linewidth]{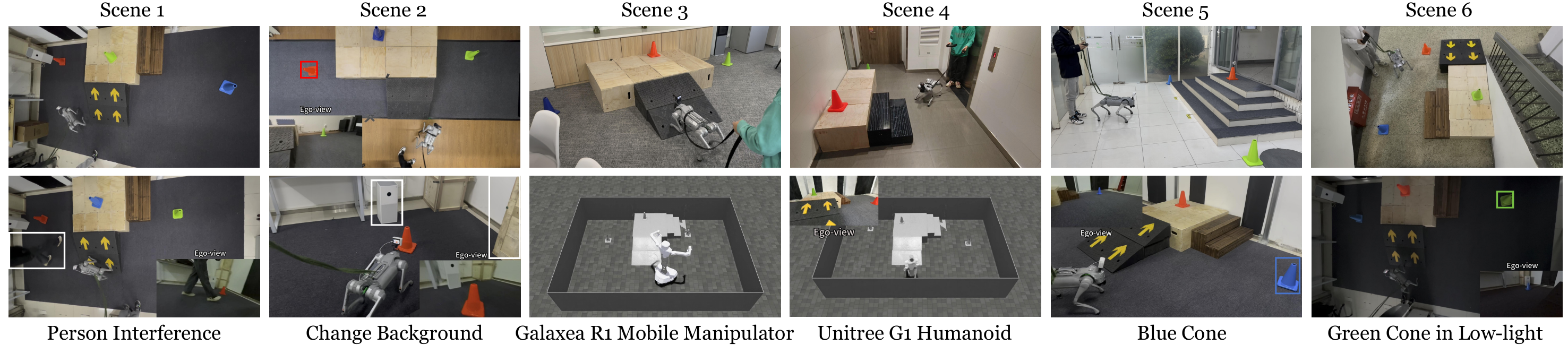}
    \vspace{-5mm}
    \caption{Diverse experiment settings including scenes, conditions, and robot types (better viewed when zoomed in).}
    \vspace{-6mm}
    \label{fig:quali_exp}
\end{figure*}   

\section{Related Works}
\label{sec:related_works}
\subsection{Sim-to-Real Policy Transfer} 
Transferring reinforcement learning (RL) policies trained in simulation to the real world remains a major challenge due to substantial domain gaps. Traditional simulator-based methods, such as domain randomization\cite{gensim, dynamic_random, domain} and system identification\cite{dynamic_motor, agile_loco}, aim to reduce the \textit{Sim-to-Real} gap by aligning simulations with physical setups. Recent work\cite{layout, semantically} proposes leveraging conditional generative models to augment visual observations for more robust agent training. LucidSim\cite{lucidsim}, for example, incorporates RGB color perception into the \textit{Sim-to-Real} pipeline to learn low-level visual parkour by generating and augmenting image background. However, these approaches remain constrained by conventional simulators, which fail to capture the full breadth of real-world physics and visual realism necessary for high-level policy training and real-world deployment.

\vspace{-0.5mm}
\subsection{Real-to-Sim Scene Transfer} 
\vspace{-0.5mm}
Recently, advances in scene representation and reconstruction such as Neural Radiance Fields (NeRF)\cite{nerf} and 3D Gaussian Splatting (3DGS)\cite{3dgs} have facilitated the creation of high-fidelity digital twins that closely replicate real-world environments for \textit{Real-to-Sim} scene transfer. For instance, NeRF2Real\cite{nerf2real} integrates NeRF and mesh into a simulation for vision-based bipedal locomotion policy training, but struggles to achieve real-time rendering, terrain climbing, and environment exploration, thus limiting their practical application. Meanwhile, RialTo\cite{reconciling} augments digital twins with articulated USD representations to enable manipulative interactions. More recent works employ 3DGS to generate realistic simulations for both robot manipulation\cite{robogs, robogsim, rl-gsbrige} and navigation\cite{quach2024gaussian,vid2sim}. In contrast, \themodel is designed to produce photorealistic and physically interactive “digital twin” environments specifically tailored for ego-centric visual locomotion learning.

\section{Task definition}

\begin{figure}[t]
    \centering
    \vspace{-2mm}
    \includegraphics[width=0.9\linewidth]{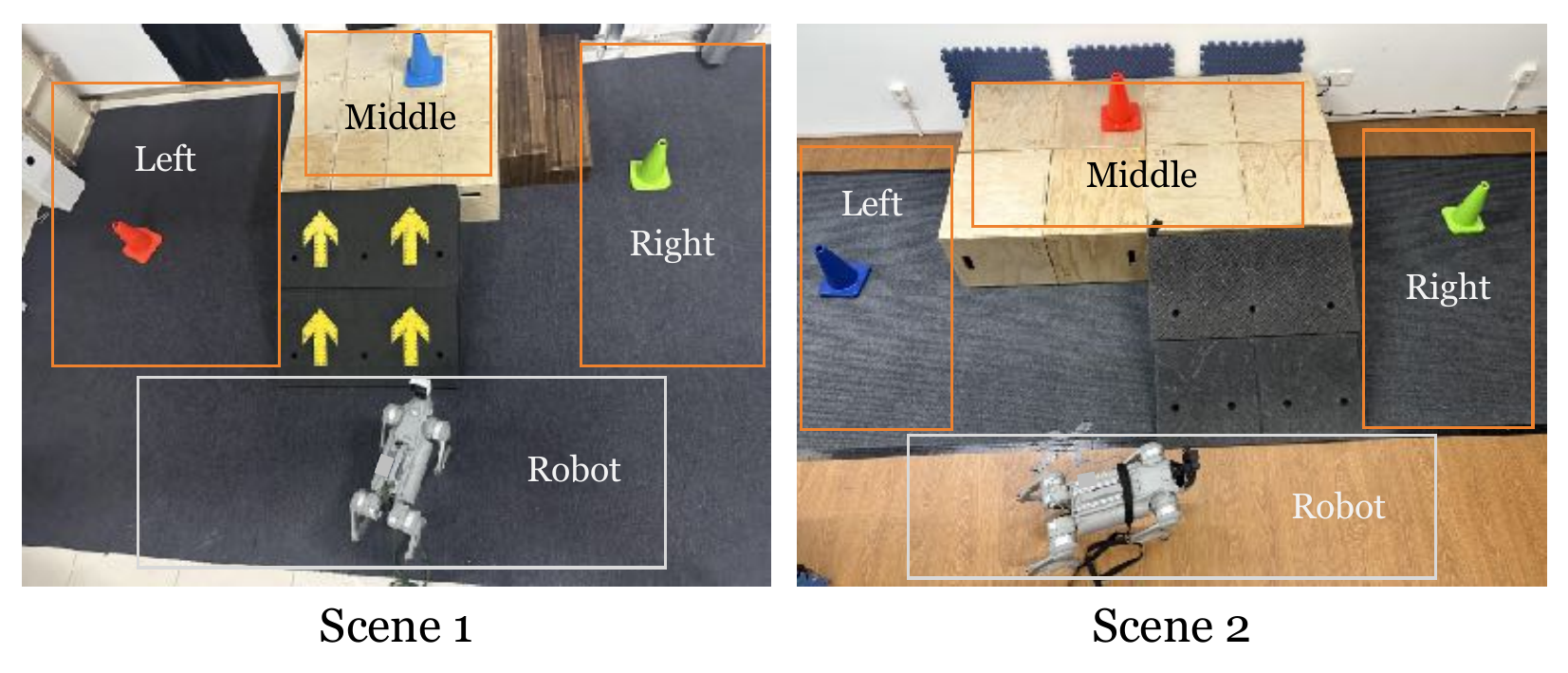}
    \vspace{-2mm}
    \caption{\textbf{Task definition.} The robot aims to reach the target cone of a specified color in a specified spatial location, relying solely on \textit{ego-view} RGB observation and proprioception.}
    \vspace{-5mm}
    \label{fig:task_def}
\end{figure}

Our designed task requires both high-level understanding and navigation, as well as low-level motion control. As shown in~\autoref{fig:task_def}, the task is defined as reaching the target cone of a specified color within a limited time horizon. Each scene features a terrain in the center composed of a high table and a slope or stair connecting the ground to the platform.  
\label{sec:task}

\begin{figure*}[ht]
    \centering
    \includegraphics[width=0.95\linewidth]{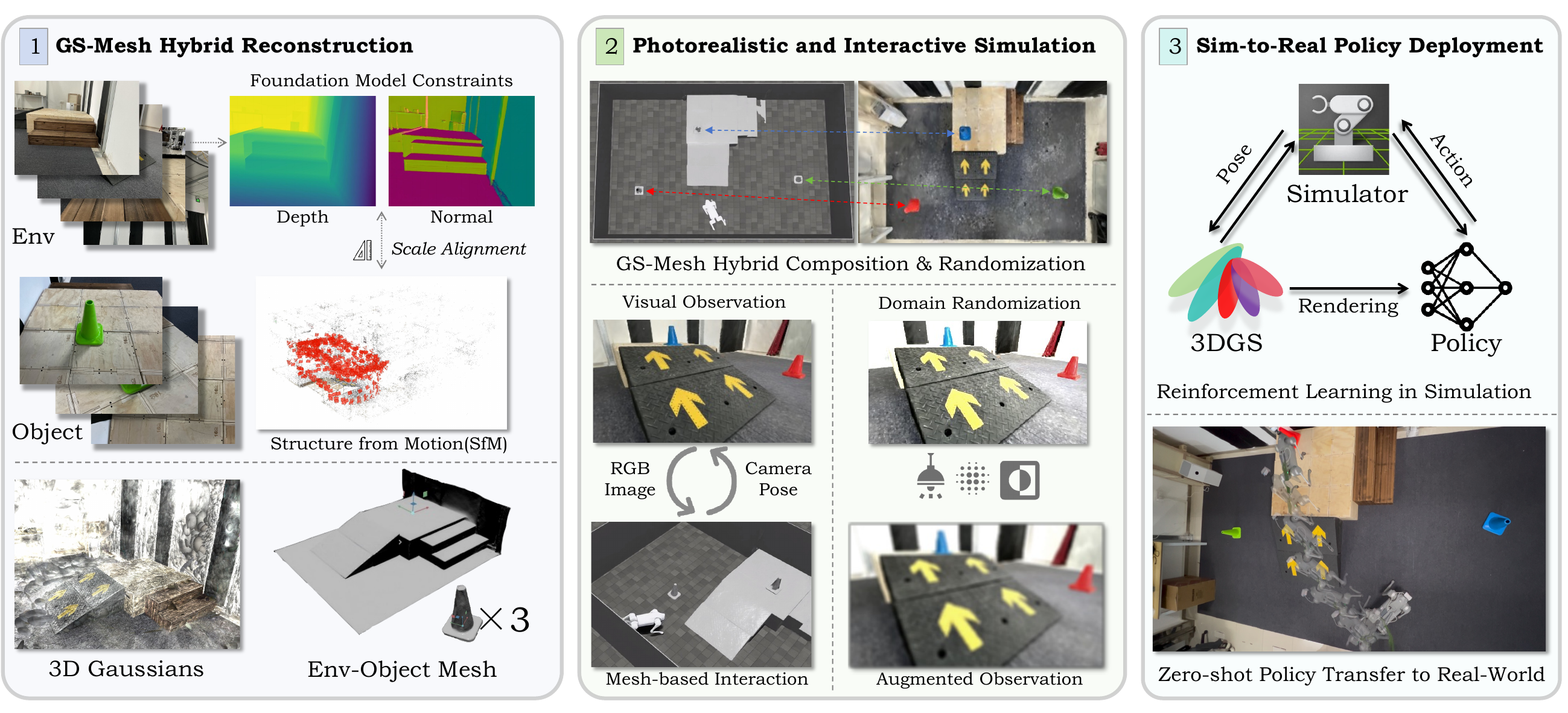}
    \vspace{-2mm}
    \caption{\textbf{VR-Robo \textit{real-to-sim-to-real} framework.} We build a realistic and interactive simulation environment with GS-mesh hybrid representation and occlusion-aware composition \& randomization for policy training. Finally, we zero-shot transfer the RL policy trained in simulation into the real robot for ego-centric navigation and visual locomotion.}
    \vspace{-6.5mm}
    \label{fig:framework}
\end{figure*}

At the start of each trial, the robot's initial position is sampled from $\mathbf{p}_\text{robot} \sim \mathcal{U}(\mathbb{P}_\text{robot})$, and its orientation is similarly randomized as $\theta_\text{robot} \sim \mathcal{U}(\mathbb{\theta}_\text{robot})$. Each scene contains three cones with identical shapes and textures but different colors $\mathcal{C}=\{\text{red, green, blue}\}$. These cones are placed in three designated regions: left, middle, and right, with one cone per region at a random position $\mathbf{p}_\text{cone} \sim \mathcal{U}(\mathbb{P}_\text{cone}^\text{region})$.  
Since the robot may not initially face the target cone, it must explore the environment to identify and locate the correct cone. Additionally, if the target cone is placed atop the table, the robot must navigate up the slope to access the platform.  

\section{A Real-to-Sim-to-Real System for Visual Locomotion}
\label{sec:real2sim2real}
\subsection{Geometry-Consistent Reconstruction}
\label{subsec:real2sim}
Given multi-view RGB images of a scene with corresponding camera poses from COLMAP\cite{colmap}, we aim to generate a photorealistic and physically interactive simulation environment for agent policy training. We decoupledly reconstruct the environment and the object, and then compose them into a single Gaussian scene for policy training.

\noindent \textbf{Gaussian-based 3D Scene Representation.} 3D Gaussian Splatting (3DGS) represents the scene as a set of Gaussian primitives as follows:
\vspace{-1mm}
\begin{equation}
    \mathcal{G}_i(\mathbf{x})=\exp \left(-\frac{1}{2}\left(\mathbf{x}-\mu_i\right)^T \Sigma_i^{-1}\left(\mathbf{x}-\mu_i\right)\right),
\end{equation}
During optimization, $\Sigma_i$ is reparametrized using a scaling matrix $S_i \in \mathbb{R}^3$ and a rotation matrix $R_i \in \mathbb{R}^{3 \times 3}$ as $\Sigma_i=R_i S_i S_i^T R_i^T$. The color value of a pixel $\mathbf{c}(x)$ can be rendered through a volumetric alpha-blending\cite{3dgs} process:
\vspace{-1mm}
\begin{equation}
    \mathbf{c}(x)=\sum_{i \in N} T_i c_i \alpha_i(\mathbf{x}), \quad T_i=\prod_{i=1}^{i-1}\left(1-\alpha_i(\mathbf{x})\right).
    \vspace{-1mm}
\end{equation}
where $\alpha_i(\mathbf{x})=o_i \mathcal{G}_i(\mathbf{x})$ denotes the alpha of the Gaussian $\mathcal{G}_i$ at $\mathbf{x} \in \mathbb{R}^3$. Meanwhile, the view-dependent color $c_i$ of $\mathcal{G}_i$ is represented by its spherical harmonics $(\mathrm{SH})$ coefficients.

\vspace{1mm}
\noindent \textbf{Flattening 3D Gaussians for Geometric Modeling.} Inspired by\cite{2dgs}, we flatten the 3D Gaussian ellipsoid with covariance $\Sigma_i=R_i S_i S_i^T R_i^T$ into 2D flat planes to enhance the scene geometry modeling by minimizing its shortest axis scale $S_i=$ $\operatorname{diag}\left(s_1, s_2, s_3\right)$:
\vspace{-1mm}
\begin{equation}
    \mathcal{L}_{\mathtt{scale}}=\frac{1}{N} \sum_{i \in N}\left\|\min \left(s_1, s_2, s_3\right)\right\|,
    \vspace{-1mm}
\end{equation}
where $N$ denotes the total number of planar-based Gaussian splats. Following PGSR\cite{pgsr}, we utilize the plane representation to render both the plane-to-camera distance map $\mathcal{D}$ and the normal map $\mathcal{N}$, then convert them into unbiased depth maps by intersecting rays with the corresponding planes as:
\begin{equation}
    \boldsymbol{D}(p)_\mathtt{render}=\frac{\mathcal{D}}{\mathcal{N}(p)K^{-1} \tilde{p}}.
    \vspace{-1mm}
\end{equation}
where $p$ is the 2D position on the image plane. $\tilde{p}$ denotes the homogeneous coordinate of $p$, and $K$ is the camera intrinsic.

\vspace{1mm}
\noindent \textbf{Foundation Model Geometric Prior Constraints.} In texture-less regions (e.g., ground and walls), photometric loss tend to be insufficient. We employ an off-the-shelf monocular depth estimator\cite{da2} to provide dense depth prior. We address the inherent scale ambiguity between the estimated depths and the actual scene geometry by comparing them to sparse Structure-from-Motion (SfM) points, following\cite{dn-splatter}. Specifically, we align the scale of the monocular depth map $\boldsymbol{D}_\mathtt{mono}$ with the SfM points projected depth map $\boldsymbol{D}_\mathtt{sfm}$ using linear regression:
    \vspace{-1mm}
\begin{equation}
    \hat{\boldsymbol{s}}, \hat{\boldsymbol{t}}=\underset{\boldsymbol{s}, \boldsymbol{t}}{\arg \min } \sum_{p \in \boldsymbol{D}_\mathtt{sfm}}\left\|\left(\boldsymbol{s} * \boldsymbol{D}(p)_{\mathtt{mono}}+ \boldsymbol{t} \right)-\boldsymbol{D}(p)_{\mathtt{sfm}}\right\|_2^2.
        \vspace{-1mm}
\end{equation}

Finally, we use the re-scaled depth map $\hat{\boldsymbol{s}} \times \boldsymbol{D}(p)_{\mathtt{mono}} + \hat{\boldsymbol{t}}$ to regularize the rendered depth map $\boldsymbol{D}(p)_\mathtt{render}$. Similarly, we adopt an off-the-shelf monocular normal estimator\cite{dsine} to provide dense normal regularization to the rendered normal map $\mathcal{N}_\mathtt{render}$ for accurate geometric modeling.

\vspace{1mm}
\noindent \textbf{Multi-view Consistency Constraints.} Inspired by\cite{massively,geo-neus}, a patch-based normalized cross-correlation (NCC) loss is applied between two gray renders $\{\mathbf{I_\mathtt{render}}, \hat{\mathbf{I}}_\mathtt{render}\}$ to force the multi-view photometric consistency: 
    \vspace{-1mm}
\begin{equation}
    \mathcal{L}_{\mathtt{mv}} = \frac{1}{\|\mathcal{P}\|} \sum_{\mathbf{p} \in \mathcal{P}} \sum_{p \in \mathbf{p}} ( 1 - NCC(\hat{\mathbf{I}}(\mathcal{H} p), \mathbf{I}(p))).
    \vspace{-1mm}
\end{equation}
where $\mathcal{P}$ is the set of all patches extracted from $\mathbf{I}_\mathtt{render}$ and $\mathcal{H}$ is the homography matrix bwteen the two frames.

\subsection{Building Realistic and Interactive Simulation}
\label{sec:hybrid-simulation}
To enable the agent-environment interaction, we integrate a GS-mesh hybrid scene representation into the Isaac Sim with coordinate alignment. We further leverage agent-object randomization and occlusion-aware scene composition to advance robust and generalizable visual policy training.

\vspace{1mm}
\noindent \textbf{GS-mesh Hybrid Representation.} \noindent 1) The \textbf{\textit{Gaussian}} representation generates photorealistic visual observations from the robot’s ego-centric viewpoints. We first align the intrinsic parameters between \textit{Sim} and \textit{Real} by calibrating the robot camera’s focal length and distortion parameters. We then obtain the camera extrinsic within the simulation coordinate system by retrieving the ego-view position and quaternion-based orientation from Issac Sim.
\noindent 2) The \textbf{\textit{Mesh}} representation facilitates physical interaction and precise collision detection. We render the depth map for each input view and leverage a Truncated Signed Distance Function (TSDF)\cite{tsdf} fusion algorithm to build the corresponding TSDF field. We then extract the mesh from this TSDF field to enable physically accurate interactions within the simulation.

\vspace{1mm}
\noindent \textbf{Coordinate Alignment.} To align the reconstructed COLMAP coordinate system with the Isaac Sim environment, we first compute the homogeneous transformation matrix \( T_{\mathtt{homo}} \in \mathbb{R}^{4 \times 4} \) by manually matching four non-coplanar points. 

\begin{figure}[ht]
    \centering
    \vspace{-4mm}
    \includegraphics[width=0.75\linewidth]{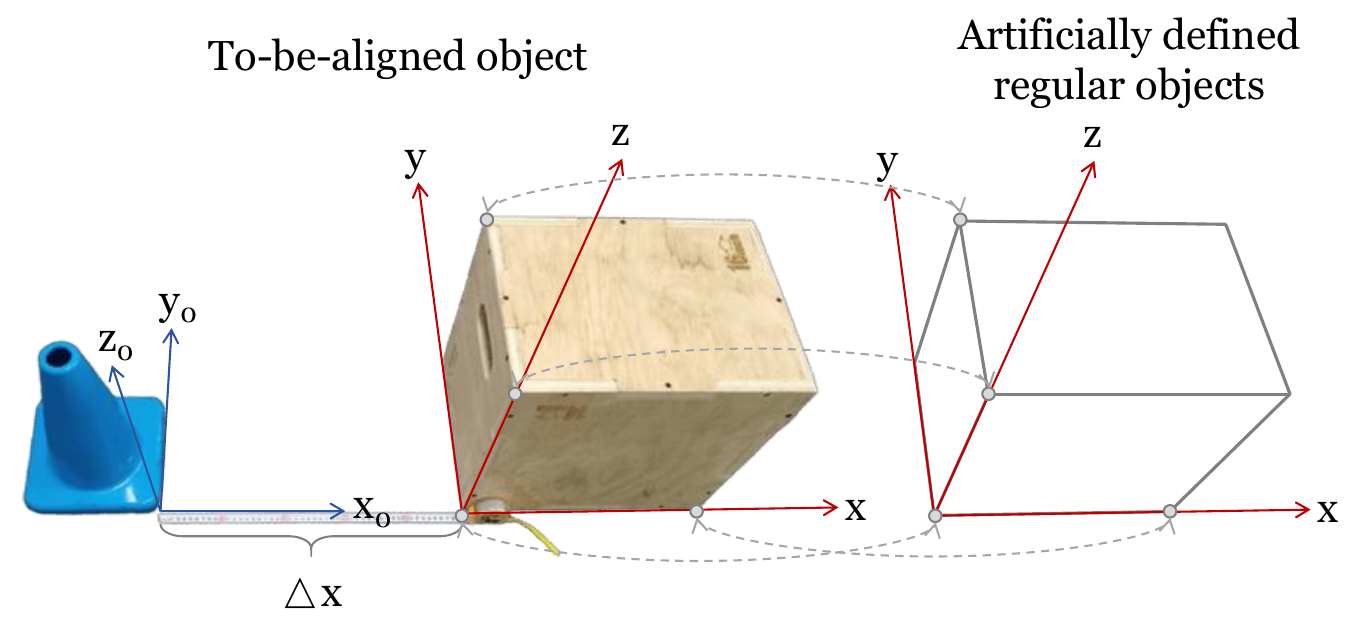}
    \vspace{-2mm}
    \caption{\textbf{Coordinate alignment.} We first compute the transformation for a regular reference object and then apply this transformation to the irregular object of interest. }
    \vspace{-3mm}
    \label{fig:4pcs}
\end{figure}

Specifically, as shown in \autoref{fig:4pcs}, we use objects with regular shapes to transform coordinate systems. For instance, to obtain the transformation matrix for the blue cone, we first manually define a rectangular block \( \mathcal{B} \) with the correct coordinate system and size as the reference object. Then, using the four-point registration method, we align the block's coordinates with the reconstructed COLMAP system. The transformation matrix \( T_{\text{block}} \) is determined as:
\vspace{-2mm}
\begin{equation}
    T_{\text{block}} = \arg\min_T \sum_{i=1}^{4} \| T \cdot \mathbf{p}_{i}^{\mathcal{B}} - \mathbf{p}_{i}^{\text{COLMAP}} \|^2,
    \vspace{-2mm}
\end{equation}

Finally, the transformation matrix for the cone, \( T_{\text{cone}} \), is determined by translating the cone's position relative to the aligned rectangular block:
\vspace{-2mm}
\begin{equation}
    T_{\text{cone}} = T_{\text{block}} \cdot \Delta \mathbf{p}_{\text{cone}}.
    \vspace{-1mm}
\end{equation}

\noindent \textbf{Gaussian Attributes Adjustment.}
We first decompose $T_{\mathtt {homo }}$ into its rotation component $R_{\mathtt {homo }} \in \mathbb{R}^{3 \times 3} $, translation vector $t_{\mathtt {homo }} \in \mathbb{R}^3$, and scale factor $s_{\mathtt {homo }} \in \mathbb{R} $. In addition, the 3D covariance matrix $\Sigma$ of the Gaussian points is parameterized by a scaling matrix $S \in \mathbb{R}^3$ and a rotation matrix $R \in \mathbb{R}^{3 \times 3}$, such that $\Sigma=R S S^T R^T$. The mean $\mu$, scaling $S$ and rotation $R$ are adjusted as follows:
\vspace{-2mm}
\begin{equation}
    \mu^{\prime} = R_\mathtt{homo} \mu + t_\mathtt{homo},
    \vspace{-1mm}
\end{equation}
\begin{equation}
    S^{\prime} = S + log(s_\mathtt{homo}),
    \vspace{-1mm}
\end{equation}
\begin{equation}
    R_\mathtt{norm} = \frac{R_\mathtt{homo}}{s_\mathtt{homo}} \quad \Sigma^{\prime} =  R_\mathtt{norm} \Sigma R_\mathtt{norm}^{T},
    \vspace{-1mm}
\end{equation}

\begin{figure*}[ht]
    \centering
    \vspace{2mm}
    \includegraphics[width=0.9\linewidth]{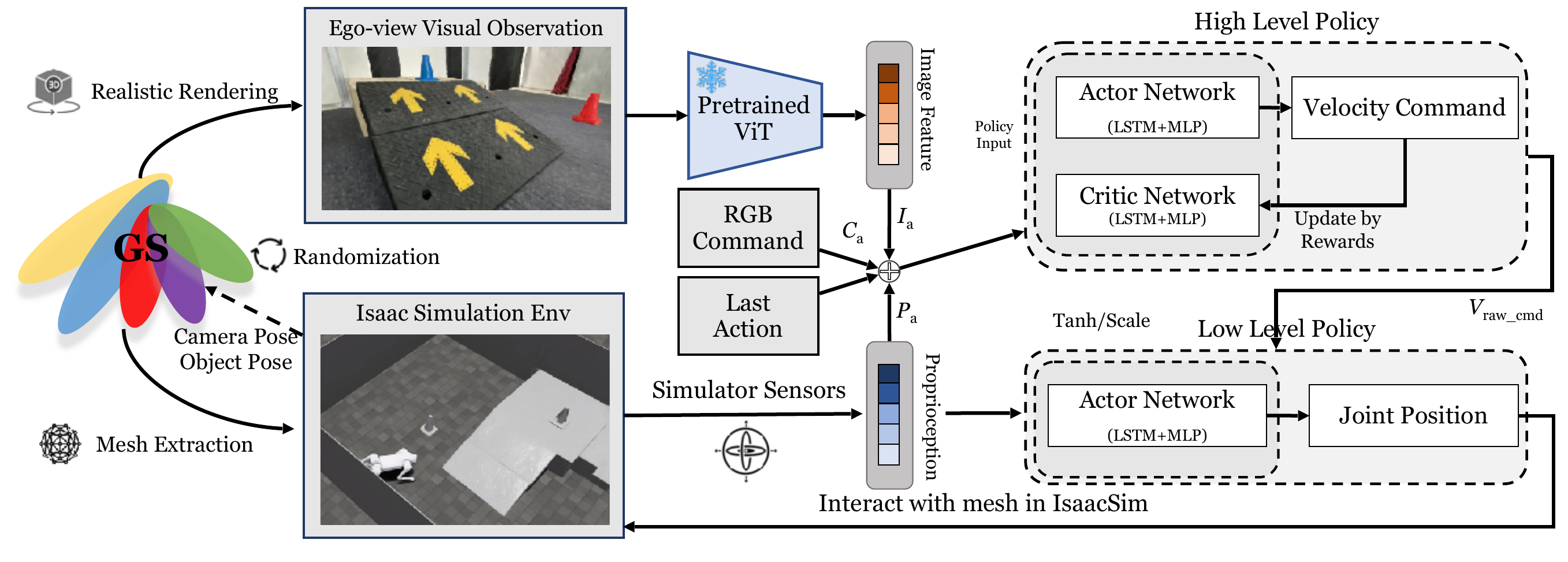}
    \vspace{-3mm}
    \caption{\textbf{RL policy training in the reconstructed simulation environment.} The agent leverages the ego-view GS photorealistic rendering as visual observations and interacts with the GS-extracted mesh in the Issac Sim environment.}
    \vspace{-6mm}
    \label{fig:rl}
\end{figure*}

Since the spherical harmonics (SHs) for the 3D Gaussians are stored in world space (i.e., the COLMAP coordinate system), view-dependent colors change when the Gaussians rotate. To accommodate different rotations, we first extract the Euler angles $\alpha, \beta, \gamma$ from the rotation matrix $R_\mathtt{homo}$ and construct the corresponding Wigner D-matrix $D$. We then apply $D$ to rotate the SH coefficients. Formally, each band $\ell$ of the SH coefficients (of length $2 \ell+1$ ) transforms as:
\begin{equation}
    \mathbf{C}^{(\ell)^{\prime}}=D_{\ell}(\alpha,-\beta, \gamma) \mathbf{C}^{(\ell)}.
    \vspace{-2mm}
\end{equation}
where $\mathbf{C}^{(\ell)}$ is the vector of the spherical-harmonic coefficients for degree $\ell$ and $D_{\ell}$ is the $(2 \ell+1) \times(2 \ell+1)$ Wigner $D$-matrix.

\begin{figure}[ht]
    \centering
    \includegraphics[width=0.8\linewidth]{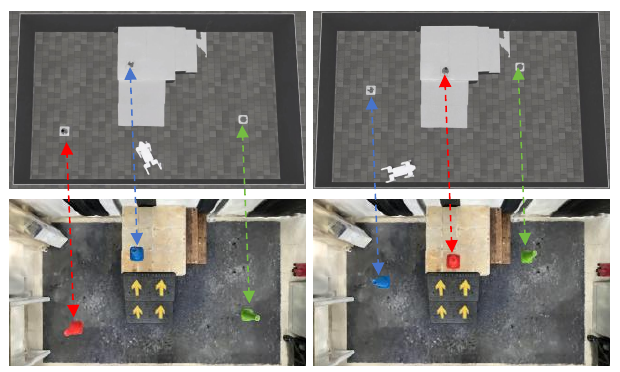}
    \vspace{-1mm}
    \caption{\textbf{Agent-object Randomization and Scene Composition.} At the beginning of each episode, we randomly sample the positions for the robot and three cones in the Isaac Sim environment (upper row). We synchronously merge the agent and object Gaussians into the environment and compose for joint rendering (lower row).}
    \vspace{-6mm}
    \label{fig:random}
\end{figure}

\vspace{1mm}
\noindent \textbf{Occlusion-Aware Composition and Randomization.} With the resulting mesh and 3D Gaussians, we employ an interactive mesh editor to obtain both the 3D bounding box and its corresponding transformation matrix $T_{\mathtt {bbox }} \in \mathbb{R}^{4 \times 4}$. Since the object mesh and the 3D Gaussians share the same COLMAP coordinate system, we use the mesh bounding box to crop the object Gaussians and merge them into the environment coordinate space according to the transformation below:
\vspace{-1mm}
\begin{equation}
    T_\mathtt{obj}^\mathtt{COLMAP-env} = T_\mathtt{sim}^\mathtt{COLMAP-env} \cdot T_\mathtt{COLMAP-obj}^\mathtt{sim} \cdot T_\mathtt{bbox}.
    \vspace{-1mm}
\end{equation}
As shown in~\autoref{fig:random}, the merged object Gaussians are synchronized with the object mesh in Issac and can be rendered jointly with the environment Gaussians via a volumetric alpha-blending~\cite{3dgs} process. Since they are separately reconstructed, there are no holes in the merged 3DGS scenes. By incorporating z-buffering~\cite{z-buffer}, it is impossible to see those objects that are farthest away from the ego-viewer and behind other objects. Therefore, our method can achieve occlusion-aware scene composition with accurate visibility. This randomization strategy substantially increases the environment's diversity, enabling more robust and scalable agent training.

\vspace{-1mm}
\subsection{Reinforcement Learning in Reconstructed Simulation}
\vspace{-0.5mm}
\label{subsec:sim2real}
The next stage of \themodel focuses on training a robust locomotion policy via reinforcement learning, as illustrated in \autoref{fig:rl}. This policy is designed to address diverse goal-tracking tasks under varying environmental conditions. Due to the limited onboard computational resources of real robots, we adopt a two-level asynchronous control strategy: the high-level policy operates at 5~Hz, while the low-level policy executes at 50~Hz. The high-level policy communicates with the low-level policy via velocity commands, represented as $V_{\text{cmd}} = (V_x, V_y, V_{yaw})$. Main experiments are conducted and demonstrated on the Unitree Go2 quadruped robot.

\vspace{1mm}
\noindent \textbf{Low-Level Policy.}
The first step is to train the low-level policy, similar to previous locomotion works\cite{wu2023learning, rrw}. The low-level policy takes velocity commands $V_{\text{cmd}} = (V_x, V_y, V_{\text{yaw}})$ along with robot proprioception as input, and outputs the desired joint positions. The actor network is composed of a lightweight LSTM layer combined with multiple MLP layers.

The low-level policy adopted here enables the quadruped robot to climb slopes and stairs measuring up to 15~cm in height. However, unlike previous parkour works\cite{parkour, cheng2024extreme, hoeller2024anymal}, this policy does not allow the robot to directly climb onto terrain taller than 30~cm. While existing frameworks could be extended to handle higher climbs, our approach instead focuses on teaching the robot to recognize and utilize slopes or stairs to reach high platforms. This strategy is particularly useful in real-world scenarios featuring taller obstacles, where direct climbing or jumping may be infeasible for any policy.

\vspace{1mm}
\noindent \textbf{High-Level Policy.}
We freeze the previously trained low-level policy and train the high-level policy independently. The high-level policy is trained using reinforcement learning (PPO) within our GS-mesh hybrid simulation environment.

\textbf{\textit{State Space}}:
The high-level policy actor network receives four types of inputs:
\texttt{1) RGB Image Feature ($I_a$)}: We employ a frozen, pre-trained Vision Transformer (ViT)\cite{vit} to extract features from RGB images. ViT's attention mechanism is well-suited for detecting positions and specific colors. This approach compresses the input size significantly and accelerates training.
\texttt{2) RGB Command ($C_a$)}: A RGB vector indicating the desired color of the target cone. For example, $[1, 0, 0]$ represents the red cone. 
\texttt{3) Last Action}: The action output by the policy during the previous timestep.
\texttt{4) Proprioception ($P_a$)}: This includes the robot’s base angular velocity, projected gravity, joint positions, and joint velocities.
We use an asymmetric actor-critic structure. The critic's observation includes the actor's observation (with noise removed) as well as the robot's world coordinate position and orientation, and the target cone's world coordinate position. This design improves the estimation of the value function and enhances the training process.

\textbf{\textit{Action Space}}:
The actor outputs the raw velocity command, $V_{\text{raw\_cmd}} = (V_x, V_y, V_{\text{yaw}})$. This raw command will pass through a $\tanh$ layer and scaled by the velocity range $V_{\text{max\_cmd}} = (V_{\text{max\_x}}, V_{\text{max\_y}}, V_{\text{max\_yaw}})$ to ensure safety. Note that the $\tanh$ layer and velocity range scaling are applied outside the actor network. The resulting velocity command $V_{\text{cmd}} = (V_x, V_y, V_{\text{yaw}})$ is sent to the low-level policy, enabling the robot to move. 

\textbf{\textit{Reward Design}}:
\label{subsec:reward_highlevel}
The total reward consists of two main categories: \textbf{Task Rewards} $r_T$ and \textbf{Regularization Rewards} $r_R$. $r_T$ include \texttt{Reach\_goal}, \texttt{Goal\_dis}, \texttt{Goal\_dis\_z}, and \texttt{Goal\_heading}. To be specific, the robot receives a reward if it comes close enough to the goal. 
\vspace{-1mm}
\begin{equation}
    r_\text{reach\_goal} = 
    \begin{cases} 
        R_\text{max}, & \text{if } \Vert \mathbf{p}_\text{robot} - \mathbf{p}_\text{goal} \Vert_2 \leq \epsilon, \\ 
        0, & \text{otherwise}.
    \end{cases}
    \vspace{-1mm}
\end{equation}

\texttt{Goal\_dis} is based on the change in the Euclidean distance to the goal between consecutive time steps:
\vspace{-2mm}
\begin{align}
    r_\text{goal\_dis} &= d_\text{prev} - d_\text{current}, \\
    d &= \Vert \mathbf{p}_\text{robot} - \mathbf{p}_\text{goal} \Vert_2.
    \vspace{-3mm}
\end{align}

\texttt{Goal\_dis\_z} is based on the change in the vertical (z-axis) distance to the goal:
\vspace{-3mm}
\begin{align}
    r_\text{goal\_dis\_z} &= d_{z,\text{prev}} - d_{z,\text{current}}, \\
    d_z &= \vert z_\text{robot} - z_\text{goal} \vert.
    \vspace{-3mm}
\end{align}

\texttt{Goal\_heading}: This reward encourages the robot to face the goal by minimizing the yaw angle error, defined as a linear function of the yaw difference:

\vspace{-4mm}
\begin{equation}
    r_\text{goal\_heading} = -\vert \psi_\text{robot} - \psi_\text{goal} \vert,
    \vspace{-1mm}
\end{equation}
where $\psi_\text{robot}$ is the robot's current yaw angle, $\psi_\text{goal}$ is the desired yaw angle toward the goal.

In addition, regularization rewards are incorporated to further refine the robot’s performance and ensure stable and efficient behavior. It contains \texttt{Stop\_at\_goal}, \texttt{Track\_lin\_vel}, \texttt{Track\_ang\_vel}, and \texttt{Action\_l2}.

\textit{Training Process}:
We randomly sample the positions and orientations of the robot and cones at the beginning of each episode. The robot's camera poses and cone poses are obtained from Isaac Sim and sent to the aligned and editable 3D Gaussians mentioned above to render a photorealistic image. The policy then outputs an action based on this image, which is applied in Isaac Sim to interact with the mesh.

\textit{Domain Randomization}:
\label{sec:domain_random}
We apply domain randomization to reduce the sim-to-real gap. This includes three key components:
\texttt{1) Camera Pose Noise}: Before requesting the 3D Gaussians to render an image, we add uniform noise to the camera pose extracted from Isaac Sim.
\texttt{2) Image Augmentation}: We randomly apply brightness, contrast, saturation, and hue adjustments to the image. This is followed by Gaussian blur to simulate camera blur caused by robot movement. Additionally, we randomly add Gaussian noise to the image with a small probability ($p = 0.05$).
\texttt{3) Image Delay:} We randomly apply a 0- or 1-step delay to the rendered images.

\section{Experiments}

\label{sec:experiments}
Our experiments are designed to evaluate two main aspects of \themodel: \textbf{1)} its ability to reconstruct realistic and interactive simulation environments with GS-mesh hybrid representation, supporting the locomotion policy training through both visual observations and mesh interactions; \textbf{2)} its capability to minimize the \textit{Sim-to-Real} gap when deployed to the real world. We conducted extensive experiments in both simulation and real-world settings to validate these capabilities.

\subsection{Experiment Setup}
\label{subsec:setup}

\noindent \textbf{Real-to-Sim Reconstruction.}
We reconstruct 6 distinct indoor room environments, each characterized by specific terrains. We also randomize the environments by incorporating three cones colored red, green, and blue. We use iPads or iPhones to take photos, which are easily available.

\vspace{1mm}
\noindent \textbf{Simulated Locomotion Training.}
We conduct policy training in Isaac Sim using a single NVIDIA 4090D GPU. For the low-level policy, we instantiate 4,096 quadruped robot agents in parallel and train for 80,000 iterations from scratch. In contrast, the high-level policy is trained from scratch over 8,000 iterations with 64 quadruped agents. The entire training process requires approximately three days to complete. For visual encoding, we employ the Vision Transformer model ``vit\_tiny\_patch16\_224'', omitting its final classification head. The Isaac Sim simulator and the 3DGS rasterization-based renderer communicate via a TCP network.

\vspace{1mm}
\noindent \textbf{Sim-to-Real Deployment.}
We deploy our policy on the Unitree Go2 quadruped robot, which is equipped with an NVIDIA Jetson Orin Nano (40 TOPS) as the onboard computer. We use ROS2\cite{ros2} for communication between the high-level policy, low-level policy, and the robot. Both policies run onboard. The robot receives desired joint positions from the low-level policy for PD control ($K_p = 40.0$, $K_d = 1.0$). We use an Insta360 Ace camera to capture RGB images. The camera captures images at a resolution of $320 \times 180$. After calibration, it has a horizontal field of view (FOVX) of 1.5701 radians and a vertical field of view (FOVY) of 1.0260 radians.

\begin{table*}[t] 
    \centering
    \vspace{2mm}
    \caption{Comparison and ablation experimental results in the real-world setting.}
    \vspace{-1mm}
    \label{tab:real-world}
    \setlength\tabcolsep{8pt}
    \fontsize{8}{10}\selectfont
    \begin{tabular}{p{3.5cm}<{\centering} | p{1.6cm}<{\centering} | p{1.3cm}<{\centering} p{1.3cm}<{\centering} p{1.3cm}<{\centering}|p{1.1cm}<{\centering} p{1.1cm}<{\centering} p{1.1cm}<{\centering}}
        \toprule
        \multirow{2}{*}{\textbf{Method}} & \multirow{2}{*}{\textbf{Exteroception}} &\multicolumn{3}{c|}{\textbf{Success Rate} $\uparrow$}&  \multicolumn{3}{c}{\textbf{Average Reaching Time (s)} $\downarrow$}\\
         & & Easy& Medium& Hard& Easy& Medium& Hard\\
         \midrule
        \textbf{Ours} & RGB & \textbf{100.00\%} & \textbf{93.33\%} & \textbf{100.00\%} & \textbf{4.96} & \textbf{6.28} & \textbf{9.09} \\
        Imitation Learning (IL) & RGB & 0.00\% & 0.00\% & 0.00\% & 15.00 & 15.00 & 15.00 \\ 
        SARO~\cite{saro} & RGB & 66.67\% & 26.67\% & 0.00\% & 46.49 & 57.24 & 60.00 \\
        \midrule
        Textured Mesh & RGB & 20.00\% & 6.67\% & 0.00\% & 12.90 & 14.90 & 15.00 \\
        CNN Encoder & RGB & 73.33\% & 66.67\% & 6.67\% & 9.10 & 11.41 & 14.90 \\
        w/o Domain Randomization & RGB & 53.33\% & 6.67\% & 0.00\% & 10.04 & 14.76 & 15.00 \\
         \bottomrule
    \end{tabular}
    \vspace{-4mm}
\end{table*}

\subsection{Simulation Experiments}
We conduct comparison and ablation experiments with various baselines. For comparison experiments, 

\begin{figure}[ht]
    \centering
    \includegraphics[width=0.8\linewidth]{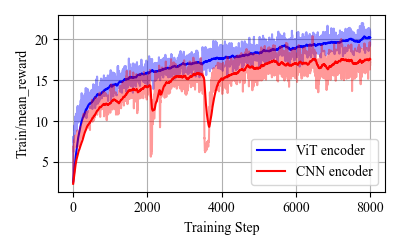}
    \vspace{-4mm}
    \caption{\textbf{Training reward comparison.} Pretrained CNN~\cite{howard2019searching} encoder struggles to extract precise semantic features, leading to unstable and worse rewards compared to ViT~\cite{vit} encoder.}
    \vspace{-3mm}
    \label{fig:encoder}
\end{figure}

\begin{figure}[ht]
    \centering
    \vspace{1mm}
    \includegraphics[width=0.77\linewidth]{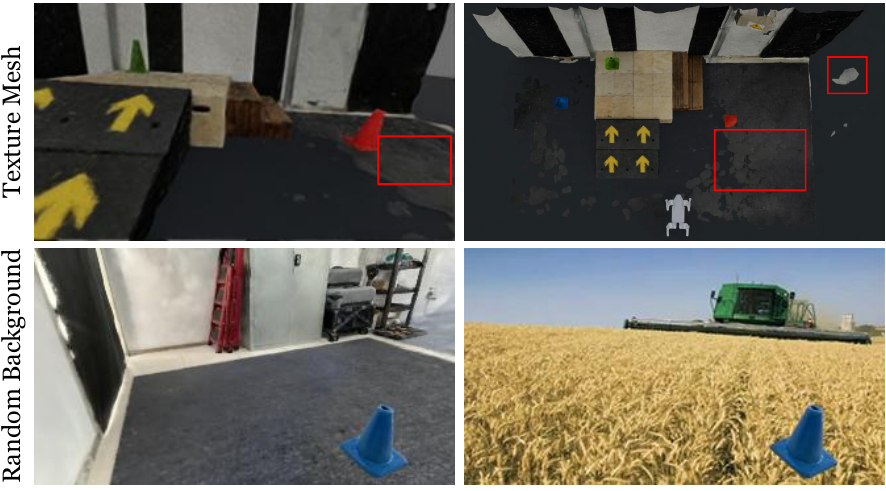}
    \vspace{-1mm}
    \caption{Illustration of textured mesh and random background.}
    \vspace{-5mm}
    \label{fig:mesh}
\end{figure}

\begin{table}[ht]
    \centering
    \caption{Comparison results in the simulation setting.}
    \vspace{-1mm}
    \label{tab:simulation_comparison}
    \setlength\tabcolsep{8pt}
    \fontsize{8}{10}\selectfont
    \begin{tabular}{p{3.0cm}<{\centering}|p{1.5cm}<{\centering}|p{1.5cm}<{\centering}}
        \toprule
        \textbf{Method} & \textbf{SR} $\uparrow$&  \textbf{ART (s)} $\downarrow$\\
         \midrule
         \textbf{Ours} & \textbf{100.00\%} & \textbf{4.94} \\
         Imitation Learning (IL) & 8.67\% & 14.01 \\
         Random Background & 43.33\% & 11.75 \\
         \bottomrule
    \end{tabular}
    \vspace{-1mm}
\end{table}

\begin{table}[h]
    \centering
    \caption{Ablation results in the simulation setting.}
    \vspace{-1mm}
    \label{tab:simulation_ablation}
    \setlength\tabcolsep{8pt}
    \fontsize{8}{10}\selectfont
    \begin{tabular}{p{3.0cm}<{\centering}|p{1.5cm}<{\centering}|p{1.5cm}<{\centering}}
        \toprule
        \textbf{Method} & \textbf{SR} $\uparrow$&  \textbf{ART (s)} $\downarrow$\\
         \midrule
         \textbf{Ours} & \textbf{100.00\%} & \textbf{4.94} \\
         Textured Mesh & 22.00\% & 12.73 \\
         CNN Encoder & 54.67\% & 10.04 \\
         \bottomrule
    \end{tabular}
    \vspace{-3mm}
\end{table}

\begin{table}[h]
    \centering
    \caption{Reconstruction quality comparison.}
    \vspace{-1mm}
    \label{tab:recon}
    \setlength\tabcolsep{8pt}
    \fontsize{8}{10}\selectfont
    \begin{tabular}{p{2.0cm}<{\centering}|p{2.0cm}<{\centering}|p{2.0cm}<{\centering}}
        \toprule
        \textbf{Method} & Planar 3DGS & Textured Mesh \\
         \midrule
         \textbf{PSNR $\uparrow$} & \textbf{29.73} & 27.73 \\
         \bottomrule
    \end{tabular}
    \vspace{-3mm}
\end{table}

\begin{itemize}
    \item \textbf{Imitation Learning:} We collect 60 different real-world trajectories via teleoperation and train the policy through supervised learning using regression optimization.
    \item \textbf{Random Background:} LucidSim~\cite{lucidsim} uses generative models to produce visual observations. Since the training and deployment code has not been open-sourced, we re-implemented it by using random images from ImageNet as the background and maintaining the target object.
\end{itemize}
\vspace{-0.5mm}
For ablation experiments,
\vspace{-0.5mm}
\begin{itemize}
    \item \textbf{Textured Mesh:} We use SuGaR\cite{sugar} to reconstruct textured mesh and support direct mesh rendering in Isaac Sim as visual observation.
    \item \textbf{CNN Encoder:} We replace ViT with a CNN~\cite{howard2019searching} encoder, and remove the last classification layer.
\end{itemize}
\vspace{-0.5mm}

We evaluated all methods in a `red cone reaching' task within Scene 1. The red cone is randomly placed in one of three regions: left (ground), middle (terrain), or right (ground). We generate 150 distinct cone placements, specified before any experiments, and apply the same setup for all methods to ensure a fair comparison.

To measure performance, we report both Success Rate (SR) and Average Reaching Time (ART). An episode is deemed successful if the robot comes within 0.25 m of the red cone at any point within a 15-second window. For successful episodes, we record the time taken to reach the cone; otherwise, we assign the maximum time of 15 seconds. The ART is then computed as the mean reaching time across all trials.


The comparison and ablation results are shown in \autoref{tab:simulation_comparison} and \autoref{tab:simulation_ablation}, respectively. Our method consistently achieves higher performance across all evaluation metrics by a large margin. In particular, the Imitation Learning (IL) baseline suffers from insufficient data samples and a lack of policy exploration. As shown in~\autoref{fig:encoder}, the CNN encoder struggles to extract precise features from the images. In this work, we reconstruct solely from RGB images. The textured mesh obtained from GS exhibits noticeable bulges in texture-less regions such as the ground and walls, as shown in~\autoref{fig:mesh} and~\autoref{tab:recon}. These artifacts degrade the rendering quality and can even hinder the robot’s movement. The quality and resolution of the mesh itself significantly affect rendering fidelity, whereas GS is less sensitive to such factors. In contrast, our method crops and utilizes only the mesh of the central terrain, and renders observation from GS, effectively avoiding these issues. The Random Background setting (the bottom-right image) loses important features specific to the original scene (the bottom-left image). These scene-specific features are crucial for the robot to effectively explore the environment.

\subsection{Real-World Experiments}

We conduct qualitative experiments under varying conditions, including 6 different scenes, various target cone colors, changes in lighting conditions, random interference, background variations, and training on the different robots, as shown in \autoref{fig:quali_exp}. A detailed demonstration is provided in the supplementary materials. These experiments highlight the robustness of our method, showcasing its adaptability to a wide range of environments and conditions.


\begin{figure}[ht]
    \centering
    \vspace{-4mm}
    \includegraphics[width=0.8\linewidth]{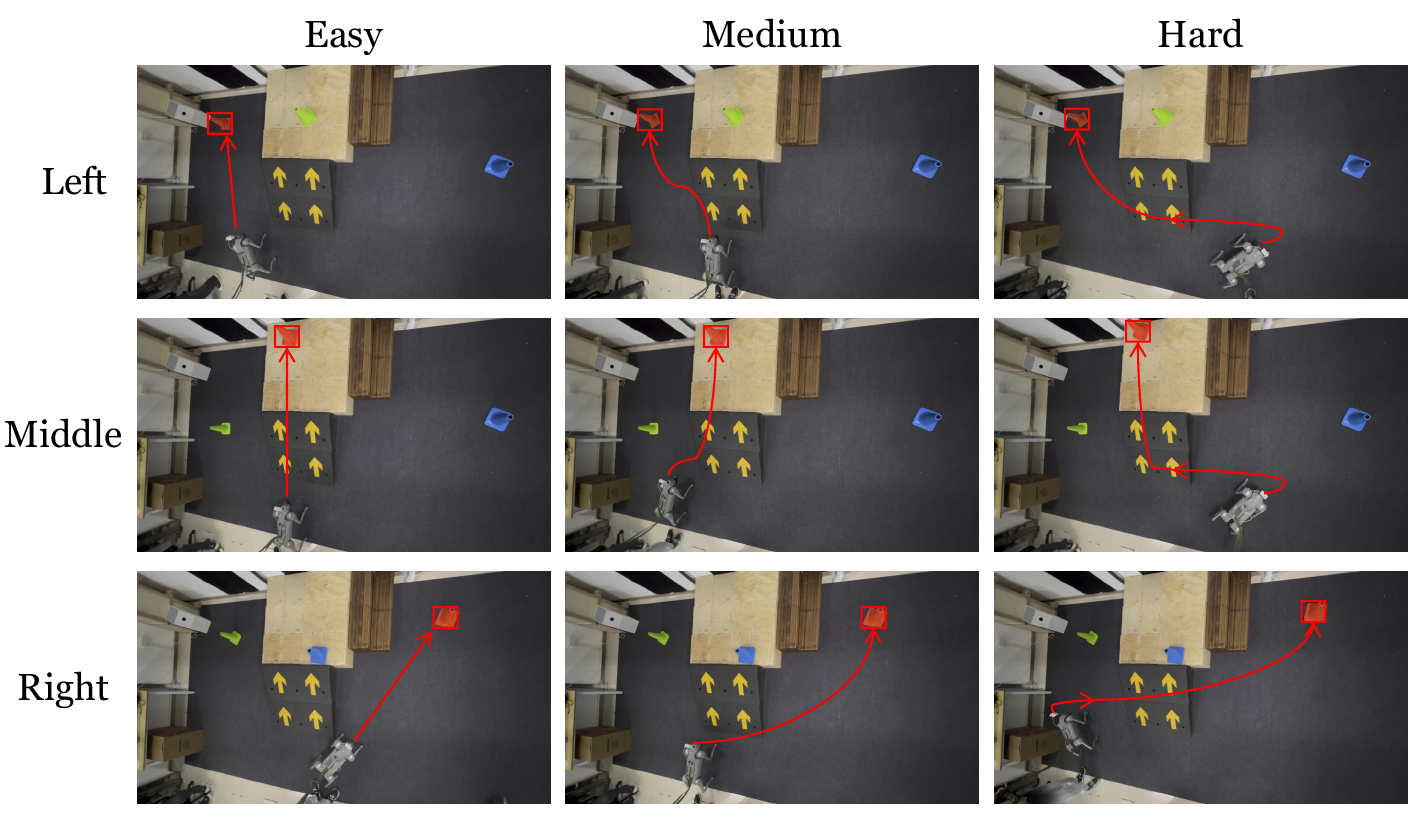}
    \vspace{-2mm}
    \caption{Tasks of different difficulties (\textit{easy}, \textit{medium}, and \textit{hard}) and different target positions (\textit{left}, \textit{middle}, and \textit{right}).}
    \vspace{-3mm}
    \label{fig:task_difficulty}
\end{figure}

We further conduct quantitative experiments, as presented in \autoref{tab:real-world}. In addition to the methods mentioned above, we conduct two additional baseline methods for real-world experiments. These methods are specifically designed in the real-world setting and cannot be directly compared in simulation:
\begin{itemize}
    \item \textbf{SARO \cite{saro}:} SARO enables the robot to navigate across 3D terrains leveraging the vision-language model (VLM).
    \item \textbf{w/o Domain Randomization:} We exclude domain randomization as described in Section~\ref{sec:domain_random}.
\end{itemize}

All the real-world quantitative experiments is in the `red cone reaching' task setting in Scene 1. We categorize the task into three levels of difficulty based on the robot's starting position: \textit{easy}, \textit{medium}, and \textit{hard}. For \textit{easy} tasks, the robot starts with the target cone directly visible. For \textit{medium} tasks, the robot needs to turn to a certain degree to see the target cone. For \textit{hard} tasks, the robot starts very far from the cone and faces a nearly opposite direction to the target cone. We randomly select 3 sets of cone positions, with each set containing one easy, one medium, and one hard task, as shown in \autoref{fig:task_difficulty}. For each task, we repeat the experiment 5 times. We then calculate the Success Rate (SR) and Average Reaching Time (ART). For SARO, the maximum time is extended to 60 seconds due to the long inference time required by the Vision-Language Model (VLM). In the real robot experiments, we consider the task successful if the robot makes contact with the target cone.

Our method achieves the highest success rates across all difficulty levels and consistently records the shortest average reaching times, demonstrating its efficiency and robustness. Note that due to inherent randomness in the real-world setting, there is a failure case in the \textit{medium} task, even though all trials succeed in the \textit{hard} task. Among the other methods, SARO achieves moderate success on easy tasks (66.67\%) but performs poorly on medium and hard tasks. This under-performance is primarily attributed to the lack of historical context and limited exploration capability in complex scenarios. If the robot cannot initially see the goal, it is unlikely to succeed.

\section{Limitations and Conclusion} 
\label{sec:conclusion}
\subsection{Limitations} 
The current environment is limited to indoor scenes with static terrains, which constrains the diversity of scenarios. Training a locomotion policy from scratch for each task takes approximately three days, posing a significant time burden. Additionally, meshes reconstructed using only RGB input often contain structural imperfections and require manual post-processing, as they lack the fidelity that could be provided by additional sensing modalities such as depth cameras or LiDAR.

\subsection{Conclusion} 
We introduce \themodel, a real-to-sim-to-real framework designed to train visual locomotion policies within a photorealistic and physically interactive simulation environment. Extensive experimental results demonstrate that our agents successfully learn robust and effective policies for challenging high-level tasks and can be zero-shot deployed to diverse real-world scenarios. In future work, we aim to extend our framework to larger-scale and more complex environments, incorporating generative models into scene reconstruction for more generalizable policy learning.
\balance




\bibliographystyle{./IEEEtran} 
\bibliography{./IEEEabrv,./IEEEexample}

\clearpage


\end{document}